\documentclass[a4paper,12pt]{article}
\usepackage{graphicx} 
\usepackage{enumitem}
\usepackage{amssymb} 
\usepackage{amsmath} 
\usepackage[left=2cm,top=3cm,right=2cm,bottom=2cm]{geometry} 
\usepackage{lastpage}
\usepackage{fancyhdr}
\usepackage{tikz}
\usetikzlibrary{arrows, positioning}
\usepackage{soul}
\usepackage{parskip}
\usepackage{hyperref}
\usepackage{xcolor}
\usepackage{soul}
\usepackage[utf8]{inputenc}
\usepackage{tcolorbox}
\usepackage[T1]{fontenc}
\usepackage{tabularx}

\usepackage[utf8]{inputenc}
\usepackage[LGR,T1]{fontenc}
\usepackage[english]{babel}

\usepackage[utf8]{inputenc}
\usepackage[T1]{fontenc}
\usepackage[labelfont=bf]{caption}
\usepackage{graphicx}
\usepackage{amsmath,amssymb}
\usepackage{enumitem}
\usepackage{tabularx}
\usepackage{tikz}
\usetikzlibrary{arrows,positioning}
\usepackage{xcolor}
\usepackage{parskip}  
\usepackage{hyperref}
\hypersetup{
    colorlinks=true,
    linkcolor=blue,
    urlcolor=blue,
    citecolor=blue
}

\usepackage{soul}
\usepackage{tcolorbox}

\title{
    \vspace{-100px}
    \textbf{\rule{\textwidth}{3pt}
    AI-Driven Contribution Evaluation and Conflict\\ Resolution: A Framework \& Design for Group Workload Investigation}

    \vspace{-15px}
    \rule{\textwidth}{2pt}
}

 \author{
     \textbf{Jakub Slapek}\\\small University of Warwick\\\small Jakub.Slapek@warwick.ac.uk \and
    \textbf{Mir Seyedebrahimi}\\\small Warwick Manufacturing Group\\\small Mir.Seyedebrahimi@warwick.ac.uk
    \and
    \textbf{Jianhua Yang}\\\small Warwick Manufacturing Group\\\small Jianhua.Yang@warwick.ac.uk
}

\date{}

\setlist[enumerate]{left=0.7cm}

\begin{document}
\cfoot{}
\maketitle



\begin{abstract}\noindent The equitable assessment of individual contribution in teams remains a persistent challenge, where conflict and disparity in workload can result in unfair performance evaluation, often requiring manual intervention – a costly and challenging process. We survey existing tool features and identify a gap in conflict resolution methods and AI-integration. To address this, we propose a framework and implementation design for a novel AI-enhanced tool that assists in dispute investigation. The framework organises heterogeneous artefacts – submissions (code, text, media), communications (chat, email), coordination records (meeting logs, tasks), peer assessments, and contextual information – into three dimensions with nine benchmarks: Contribution, Interaction, and Role. Objective measures are normalised, aggregated per dimension, and paired with inequality measures (Gini index) to surface conflict markers. A Large Language Model (LLM) architecture performs validated and contextual analysis over these measures to generate interpretable and transparent advisory judgments. We argue for feasibility under current statutory and institutional policy, and outline practical analytics (sentimental, task fidelity, word/line count etc.), bias safeguards, limitations, and practical challenges.

\vspace{10px}\noindent \textbf{Keywords:} \textit{AI-assisted assessment}, \textit{Conflict resolution in teams}, \textit{Contribution investigation}, \textit{Peer assessment}, \textit{Collaborative learning}, \textit{Learning analytics}.
\end{abstract}

\section*{[1] Introduction}

Collaborative tasks and teamwork are a cornerstone of contemporary higher-education pedagogy, cultivating skills that are difficult to develop in solitary coursework. However, where individual grades need to be determined and some participants contribute unevenly, disagreements may arise. Adjudicating these disputes is often challenging and time-consuming due to the quantity of evidence, relationship nuances, ambiguity, and many other factors. On top of this, the final verdict itself can be influenced by subjectivity and bias.

Common practice assigns grades equally among team members, which can lead to negative reactions from students, and does not reflect the typical industry experience\cite{oakley2004turning}. AutoRating systems and web-based peer-assessment platforms enable peers to report relative contributions and automate assessment based on rubrics via instruments; however, these increase administrative load, suffer from subjective popularity effects, and aid minimally in conflict resolution.

Parallel work in Learning Analytics (LA) has sought objectivity via analysis of student artefacts to derive equity metrics, measures of presence, and emergent roles. Yet, current research centres on the initial stage of feedback\cite{tempelaar2024dispositional}, and supports student agency mainly in theory – operational practices remain sparse\cite{hooshyar2023learning}.

AI/Generative AI (GenAI) offers a way to bridge these fields by contextualizing multiple sources of evidence and the semantics of human interaction under a “neutral” stance\cite{Bisoux2025HarnessingAI}. Existing methods and modern tools reserve AI for screening and semantic analysis unrelated to contribution evaluation, and GenAI for qualitative feedback and resource generation in manners that are invasive, unsuitable towards summative assessment, and lack transparency and explainability. Meanwhile, frameworks target ethics and allowance in peer assessment, but aside from advising to keep the “human-in-the-loop”, few discuss the evaluation process itself.

Hence, there is a definitive gap in both systems that provide in-depth assistance for group workload investigation, and methodology and guidance towards advisory assessment practices.

In this work, we establish a foundation for the development of a tool enhanced with AI designed to assist the investigation of disputes over individual contribution in group work. Our investigation assesses the feasibility of AI-assisted investigation in higher education under existing policy, and comprehensively surveys dispute investigation methodology and AI integration in existing tools.

We propose: (1) A conflict assessment framework suggesting three dimensions (\textit{Contribution}, \textit{Interaction}, and \textit{Role}), subdivided into nine benchmarks, signalling conflict based on categorised evidence; (2) An implementation plan instantiating our framework – a pipeline from evidence to AI-driven judgment – that extracts features over submissions, communications, and coordination records in the form of objective metrics, normalises and aggregates values, analyses inequality to raise conflict markers, and performs evaluative \textit{advisory} judgment.

\vspace{-5px}\section*{[2] Background Study}

\subsection*{[2.1] Foundation}

\cite{topping1998peer} defines Peer Assessment (PA) as “a system in which individuals evaluate the quantity, level, value, worth, quality, or achievement of their peers’ learning products or outcomes, who share a similar status”. It has been shown that peer marks correlate strongly with instructor marks\cite{Falchikov2000}, and that well-scaffolded PA cultivates students' assessment literacy and self-regulation skills – two prerequisites for fair contribution judgements in group work\cite{yang2025peer}\cite{Panadero2018}. 

This is deeply entwined with the field of Collaborative \& Co-operative Learning (CL)\cite{serc2006cooperative}\cite{panitz1999collaborative}, whose benefits have been displayed in countless studies\cite{oakley2004turning}, and whose failing mechanisms lead to an erosion of group learning and morale\cite{Slavin1996}. On the other hand, it can also be the case that assessment becomes the “Achilles heel” of learning, inhibiting the processes it is designed to enhance\cite{Boud2001PeerLearning}. We discuss governing factors through the lens of computer-assisted \textit{summative} assessment that facilitate CL.

A major driver of unfairness is Social Loafing: "students who don’t take responsibility for their own role, even if it is the smallest role in the group"\cite{Isaac2012}, often leading to an over-reliance on stronger students\cite{Lunetta1990} who resent others for gaining credit for what they perceive as their own contribution\cite{Boud2001PeerLearning}. However, whilst it has been shown that normalisation is an effective fix for generosity bias and yields measurably fairer grades, a minority of students react negatively because the algorithm feels like it over-rides their personal judgements\cite{stonewall2022development}. Therefore, beyond \textit{procedural} fairness, tools must also earn social legitimacy to balance technocracy.

Context and accuracy are major talking points from \cite{Falchikov2005Improving}. Results are reproducible from quantitative data, but more meaningful from qualitative data. The author cautions that a heavy skew in either direction can lead to test pollution (i.e., learning for the exam), and an inability to account for students’ knowledge. Increasing inter-rater reliability (that multiple markers would agree with the same evaluation) reduces instructor bias, but can be impractical due to the institutional costs of many markers. Subsequently, they suggest algorithmic co-marking and structured peer and self-assessment.

\vspace{-7px}\subsection*{[2.2] Peer Assessment Instruments}

Adding a qualitative dimension via structured psychometric tools demonstrates improvements in the fairness and accuracy of PA\cite{Ridwan2025Bias}, as well as inter-rater reliability\cite{Jonsson2007}. Moreover, studies show rubrics enhance the quality of evaluative judgment\cite{Gyamfi2021Rubrics}.

The AutoRating System\cite{brown1995autorating} derives individual marks from team marks – tackling “passenger syndrome” (unequal contribution), and promoting individual learning in team projects. Students rate their peers in qualitative categories (e.g., “Excellent”, “Unsatisfactory”), each assigned a numerical value from 0 to 100, and are “autorrated” as follows:
\begin{equation}
    \text{\textit{Final Indiv. Grade}} = (\text{\textit{Indiv. Average }} / \text{ \textit{Team Average}}) \times \text{\textit{Team Grade}}
\end{equation}
Assessment guidelines for instructors in the CL literature recommend the system\cite{oakley2004turning}, and studies promote its use for decreasing administrative load and identifying dysfunctional teams and team members (“hitchhikers” and “tutors”)\cite{kaufman2000accounting}\cite{asee_peer}. Variants include square rooting the weight factor, and constant sum allocation with a fixed number of distributable points, to which studies report some collusion risk, but decreased rating inflation\cite{goldfinch1990development}\cite{goldfinch1994further}.

These ordinal judgements can be expanded upon to produce more comprehensive instruments by diversifying and segregating surveyed data: a rubric that contains items (i.e., singular survey questions) sorted into categories. CATME\cite{loughry2007development} is at the forefront of peer evaluation instrument research, with a rubric containing 87 items sorted into 29 types of team-member contributions, producing scores in five categories: \textit{Contributing to the Team’s Work}, \textit{Interacting with Teammates}, \textit{Keeping the Team on Track}, \textit{Expecting Quality}, and \textit{Having Relevant Knowledge, Skills, and Abilities}. They choose to augment this Likert-scale format by introducing behaviour-level descriptions and reducing the number of items to decrease rating time and cognitive load\cite{ohland2012comprehensive}.

The main issue with instruments constructed from PA surveys is \textit{bias}, which is difficult to avoid and account for\cite{brindley1998peer}. Studies show that bias can be observed in strategic behaviours like reciprocity to inflate reports\cite{li2024reciprocity}, and methods such as bias-correction normalisation (that uses the ratio between the ratings received and given) help mitigate over-generosity and creative accounting, but don’t remove the problem outright\cite{li2001refinements}.

\vspace{-7px}\subsection*{[2.3] Learning Analytics}

Learning Analytics (LA) is a growing field, fueled by the ’Data Explosion’, rise of ’Big Data’, and the need for reform in the efficiency and quality of higher education\cite{Long2011Penetrating}. It governs the measurement, collection and analysis of learner data, and contains a subfield called Collaboration/Collaborative LA (CLA) that tackles group learning\cite{MartinezMaldonado2021Collaboration}.
Frameworks that guide tool development cover stakeholders, limitations, and instruments\cite{Greller2012Translating}, as well as ethical principles such as robustness and explainability\cite{Slade2013Learning}. Adjacent work in formative assessment discusses assessment targeting \textit{behaviours} over products\cite{arthars2024formative}, and how to adopt existing frameworks with transparency-based principles like ACAD\cite{goodyear2021acad}.

The trend in LA is to combine subjective/qualitative metrics (PA instruments) with objective/quantitative metrics (student artefacts). \cite{katsenos2025assessing} employs Peer Contribution and Digital Presence scores by dividing peer ranks by maximum potential score per task and number of digital accesses. Others combine objective Git metrics (such as ‘relative share’) with questionnaires\cite{mitra2024analyzing} or pruned CATME scores\cite{Buffardi2020Assessing}, along with contribution ‘equity’ using inequity measures like Gini\cite{hundhausen2023combining} as an early warning system for disparity. 

Different approaches analyse student-subtask bipartite networks to identify emergent roles to judge the quantity and heterogeneity of individual contribution\cite{feng2025analyzing}, as well as Multimodal Learning Analytics that appropriates sensory data to perform epistemic network analysis, introducing behaviour-level encoding not present in typical instruments\cite{sung2024beyond}\cite{kiafar2025mena}\cite{zhao2024ena_endusers}. The latter especially is being increasingly enhanced by AI\cite{Mohammadi2025AIinMMLA}.

Overall, the above investigations call for improved instruments to capture peer evaluation that counters popularity inflation “Halo” effects, and objective metrics that capture intangible contributions.

\vspace{-7px}\subsection*{[2.4] Artificial Intelligence}

Reviews on AI in Assessment agree precedence needs to be placed on ethical considerations, bias mitigation techniques, and above all, human oversight\cite{Zhao2024AIassistedAssessment}\cite{zhao2024generative}\cite{Liang2025EarlyImpactAIHE}\cite{nikolopoulou2024generative}.  Ethical frameworks like FATE have been well-established for years\cite{dwork2011fairness}, inspiring projects like Microsoft’s Fairlearn\cite{fairlearn}, and some work exists on frameworks for integrating AI with assessment methods\cite{Perkins2024AIAssessment}. But when it comes to \textit{evaluation} and enhancing decision-making for instructors – the level at which human oversight should occur and how integration with AI can improve the process – guidance is sparse.

\cite{Darvishi_2022} proposes a system to categorise AI use for evaluation. They gauge the “trustworthiness” of each team member via probabilistic models along with analytics and the cosine similarity to semantic encodings of feedback sourced from a web-based platform. \cite{Topping2025} extends the framework to cover the affordances of AI in peer assessment. They outline a lack of transparency in current AI applications and a focus on timesaving for instructors over added value. A parallel study using GenAI used prompts with students’ work and a 1-5 score peer assessment rubric, followed by manually extractive qualitative feedback and grades from the chat transcript. Feedback was of higher quality, but sometimes contradictory and incorrect, and its consistently higher scoring implied an inherent “kindness bias”\cite{usher2025generative}.

\section*{[3] Existing Tool Survey}

Assessment improves with multiple raters, detailed rubrics, and moderation meetings to calibrate standards\cite{Watari2022Effect}\cite{Tong2023Raters}\cite{Jonsson2007}. However, which methods are actually used in practice? This section surveys the most widely used systems for assessment in group learning tasks that have a notion of contribution. Specifically, we focus on web-based tools with research output that have had the largest adoption in higher education, and delve into features that may be helpful towards dispute resolution. We look at eleven (11) items: CATME\cite{ohland2012comprehensive}, SPARK (PLUS)\cite{freeman2002spark}, WebPA\cite{Loddington2009WebPA}, RiPPLE\cite{Khosravi2019RiPPLE}, Peerceptiv (previously SWoRD)\cite{Cho2007Scaffolded}, TEAMMATES\cite{teammates}, Buddycheck\cite{buddycheck}, iPeer\cite{ipeer2006}, Eduflow\cite{eduflow}, InteDashboard\cite{intedashboard}, and FeedbackFruits\cite{feedbackfruits}.


\vspace{-5px}\begin{figure}[htbp]
    \centering
    \includegraphics[width=0.9\textwidth]{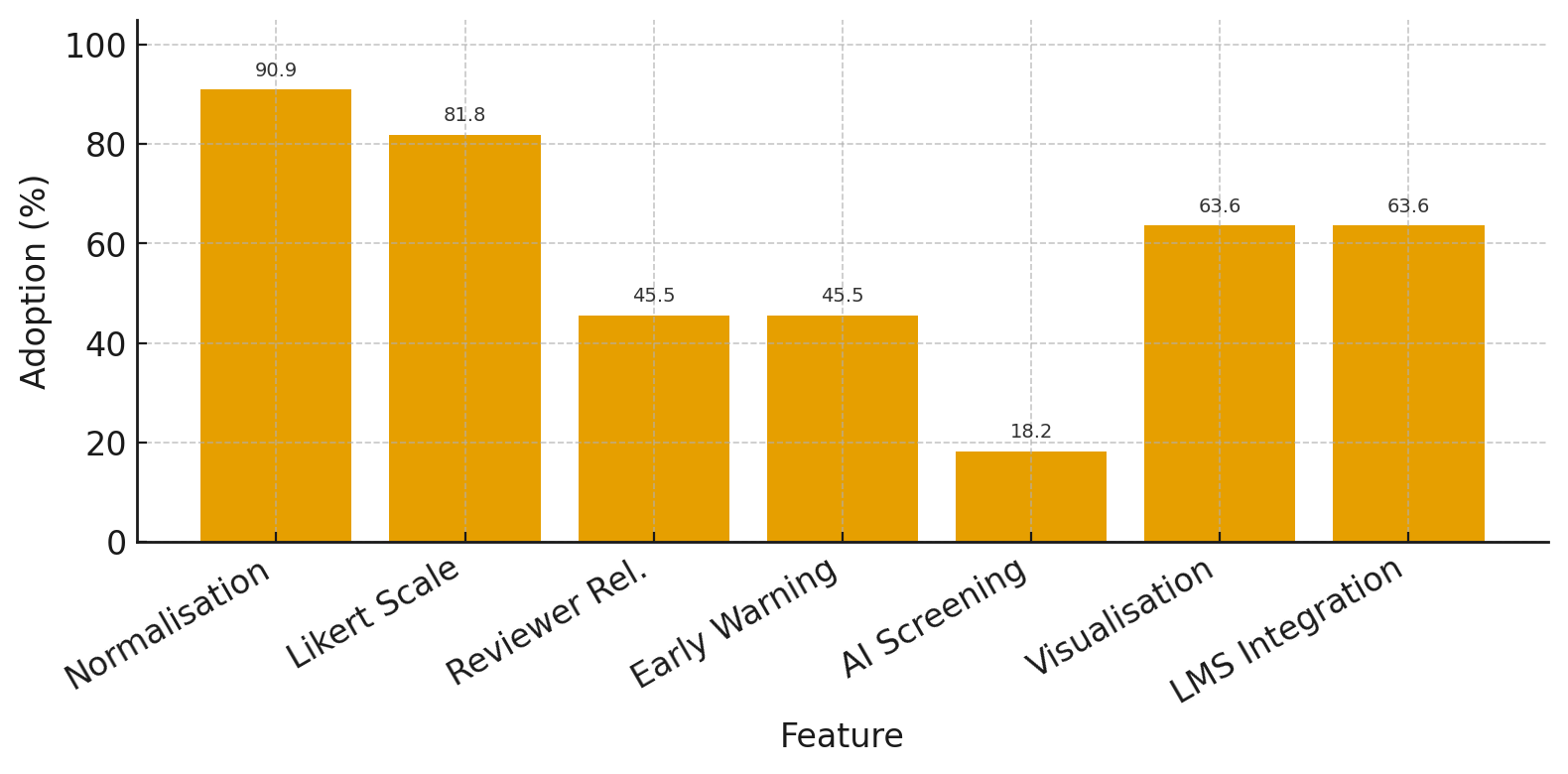}\vspace{-5px}
    \caption{Bar graph of feature adoption in contribution assessment tools.}
    \label{fig:feature-adoption}
\end{figure}

\vspace{-5px}Of the tools listed above, the majority (91\%) used an AutoRating-derived normalisation to determine scores based on peer evaluations, naming the metric “relative performance factor” or “group factor”. Variations include square-rooting the weight factor (FeedbackFruits), and constant sum allocation (iPeer). Additionally, 82\% constructed rubrics based on Likert-style surveys, with exceptions employing behaviourally anchored ‘levels of mastery’ (CATME-B) and custom multimodal rubrics (FeedbackFruits).

Notions of reviewer reliability were seen in 45\% of tools. RiPPLE and Peerceptiv used probabilistic models produced by iterative algorithms to gauge member trustworthiness, with binary approval from social graphs and “helpfulness” scores respectively. Eduflow created a “review quality index” to allocate partners based on feedback analytics, whilst InteDashboard and BuddyCheck utilized the standard deviation of peer evaluation scores, penalising missing reports.

Early warning systems were seen in 45\% to flag high/low performers, cliques, conflict, etc. based on standard deviation, skew, and raw totals. 18\% used AI screening for submissions: spot-checking “suspicious” work (RiPPLE), and binary acceptance based on the instructor rubric (FeedbackFruits). 64\% displayed analytics on a dashboard or visualized metrics, such as radar diagrams and individual vs. team average comparisons. Finally, 64\% offered learning management system (LMS) integration with platforms like Moodle\cite{moodle_lms} and Canvas\cite{canvas_lms}.

Overall, AI was mainly utilised to screen submissions or determine reliability/helpfulness (though some, like RiPPLE, developed learning aids with content/resource generation). For conflict resolution, early warning systems and manual flags – such as “objection mode” (SPARK) and “Flag-to-Teacher” (FeedbackFruits) – informed instructors of conflict. However, only Peerceptiv allowed for students to add explanatory evidence (though no additional analysis is given). Thus, there is a deficiency in current tools towards AI integration and conflict investigation features.

\vspace{-10px}\section*{[4] Proposed Conflict Framework}

Prior research has introduced instruments for evaluating student performance in collaborative tasks. Yet existing approaches seldom address conflict in a comprehensive manner. Inspired by works like CATME, we introduce a framework that lays the groundwork for an instrument benchmarking items which indicate behaviours associated with conflict. Unlike prior instruments, we incorporate objective artefacts and leave benchmark metrics open to implementation-specific interpretation based on evidence type and data availability.

Specifically, we (1) categorise evidence, (2) define benchmarks and the scope of their input, and (3) report potential points of conflict that each benchmark may signiy. This framework prioritises practicality so that it may be used as a foundation for tool development, whilst maintaining a broad coverage of issues that an instructor would want to be aware of to judge individuals within a group fairly.

\begin{table}[htbp]
    \centering
    \begin{tabular}{lll}
        \hline
        \textbf{1. Submission} &
        \textbf{2. Conversation} &
        \textbf{3. Coordination} \\
        \hline
        \begin{tabular}[t]{@{}l@{}}
            \rule{0pt}{2.5ex}a. Code\\
            b. Text\\
            c. Multimodal
        \end{tabular}
        &
        \begin{tabular}[t]{@{}l@{}}
            a. Discussion Log\\
            b. Email\\
            c. Meeting Minutes
        \end{tabular}
        &
        \begin{tabular}[t]{@{}l@{}}
            a. Personal Circumstances\\
            b. Miscellaneous Artefacts\\
            c. Task Description\\
            d. Peer Assessment
        \end{tabular} \\
        \hline
    \end{tabular}
    \vspace{-4px}
    \caption{Evidence categorisation.}
\end{table}

\vspace{-5px}We label the evidence, as seen above. Note that we define "personal circumstances" here as events that occur during the assessment process that may cause absence. Due to the subjective and complex nature of personal issues, previous student experience, and academic history, we do not delve into how these may affect the performance and behaviour of a student.

\renewcommand{\arraystretch}{1.5}
\vspace{5px}\begin{table}[h!]
\centering
\begin{tabularx}{\textwidth}{|p{0.15\textwidth}|p{0.12\textwidth}|p{0.22\textwidth}|p{0.4066\textwidth}|}
\hline
\textbf{Benchmark} & \textbf{Evidence} & \textbf{Description} & \textbf{Conflicts}\\
\hline
\centerline{Quantity} & (1) & The raw amount a student contributed to the submitted work. & Low contribution as a result of social loafing and hitch-hiking. Over-centralisation where individuals "hog" tasks. \\ \hline
\centerline{Quality} & (1), (3c) & How "good" the submitted work was. This can be based on a (possibly AI-generated) rubric. & Low effort resulting in poor work, opaque, difficult-to-understand styles, duplicate work (i.e., plagiarism), lack of skill and knowledge. \\ \hline
\centerline{Relevance} & (1), (3c) & Whether the contributed work provides progress. Did the work match the task description? & Unfocused team, members not following expected learning out-comes, poor critical thinking and task understanding, avoiding im-portant tasks. \\ \hline

\end{tabularx}
\vspace{-5px}
\caption{Contribution.}
\label{tab:equal_width}
\end{table}

\newpage

\begin{table}[h!]
\centering
\begin{tabularx}{\textwidth}{|p{0.15\textwidth}|p{0.12\textwidth}|p{0.22\textwidth}|p{0.4066\textwidth}|}
\hline
\textbf{Benchmark} & \textbf{Evidence} & \textbf{Description} & \textbf{Conflicts}\\
\hline
\raggedright
\centerline{Tone} & (2), (3d) & An acceptable display of behaviour by students. & Rude communications, disagreements, unequal/cliquey treatment, negativity, and non-constructive feedback. \\ \hline
\centerline{Effectiveness} & (2) & The effectiveness of team communications, and how well information was relayed. & Lack of clarity in explanations, short and sparse messages, failure to keep the team on track, poor phrasing, misunderstanding about each other’s skills and knowledge. \\ \hline
\centerline{Presence} & (2), (3a) & Was the student present to engage with the task? & Absences (possibly due to personal circumstances), long periods of time between responses. \\ \hline

\end{tabularx}
\vspace{-5px}
\caption{Interaction.}
\label{tab:equal_width}
\end{table}

\begin{table}[h!]
\centering
\begin{tabularx}{\textwidth}{|p{0.15\textwidth}|p{0.12\textwidth}|p{0.22\textwidth}|p{0.4066\textwidth}|}
\hline
\textbf{Benchmark} & \textbf{Evidence} & \textbf{Description} & \textbf{Conflicts}\\
\hline
\raggedright
\centerline{Adherence} & (1),(2),(3c) & The adherence of students to the published deadlines and internal agreements. & Tardiness, bad planning, and not meeting deadlines. Doing the work of others, and not following plans set out by the team, over-promising. \\ \hline
\centerline{Organisation} & (1),(2),(3a),
(3b) & The level of organisation in the team. & Scattered files, version control issues, storing artefacts locally and not sharing with the team, chaotic work, and unprepared at meetings. \\ \hline
\centerline{Support} & (1),(3b),
(3d) & Administrative participation or helping the team indirectly.
 & Avoidance of planning, setting up the workspace, research, administration, etc. Lack of motiva-
tion, knowledge/skill hoarding.
 \\ \hline

\end{tabularx}
\caption{Role.}
\vspace{-5px}
\label{tab:equal_width}
\end{table}

We consider three overarching benchmark categories: \textit{Contribution}, \textit{Interaction}, and \textit{Role}. "Contribution" centres around work submitted by students for the group task that is gradable. "Interaction" governs how students communicated, and determines whether a student’s behaviour is acceptable. "Role" captures the effectiveness of a student as a team member, encompassing less concrete methods of contribution in addition to adherence to group dynamics. Within every category, we establish three benchmarks with descriptions, the types of evidence that may be used to calculate metrics, and potential points of conflict resulting from poor scores in a given benchmark.

\section*{[5] Implementation Plan}

In this section, we explore the implementation of the conflict framework proposed above to inform the development of a tool that assists instructors in adjudicating group work disputes. Our design (1) provides instructors with numerical metrics to assess student performance in team tasks objectively, (2) alerts instructors about potential points of conflict, signs of negative behaviour, and group dynamics, and (3) offers holistic evidence-driven judgements paired with subjective interpretations of data and context to mirror expert opinion. To account for the varied availability of evidence, we propose stages to our architecture, whose investigative power scales with the diversity of its knowledge base.


\begin{figure}[htbp]
    \centering
    \includegraphics[width=0.5\textwidth]{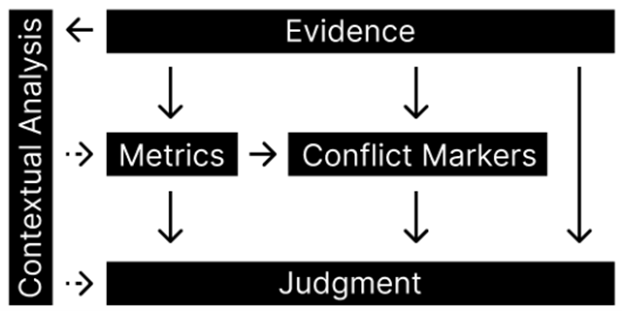}\vspace{-2px}
    \caption{High-level architecture overview.}
    \label{fig:feature-adoption}
\end{figure}

As seen above, the system accepts evidence to produce metrics and conflict markers that act as the foundation for feature inference and expert analysis. Optionally, a contextual stage can be integrated to affect calculations and conclusions. We discuss each stage in more detail below.

\subsection*{[5.1] Metrics}

Our base design calculates analytics based on data collected from (1) submitted work in the form of text, code and other multimodal output, (2) chat records, emails, and any other communication media, and (3) meeting logs, task threads, and miscellaneous artefacts. These metrics are calculated for each student and aim to objectively capture individual performance.

\begin{table}[htbp]
    \centering
    \renewcommand{\arraystretch}{1.1}
    \begin{tabular}{lll}
        \hline
        \textbf{1. Submission} &
        \textbf{2. Conversation} &
        \textbf{3. Coordination} \\
        \hline
        \begin{tabular}[t]{@{}l@{}}
            \rule{0pt}{2.5ex}a. No.\ of Commits\\
            b. Code Line Count\\
            c. Word Count\\
            d. Character Count\\
            e. Avg.\ Time Interval\\
            f. Weighted Skew\\
            g. Code Standard\\
            h. Text Complexity\\
            i. Media Workload
        \end{tabular}
        &
        \begin{tabular}[t]{@{}l@{}}
            a. Message/Send Count\\
            b. Char/Msg.\ Ratio\\
            c. Send/Receive Ratio\\
            d. Avg.\ Response Time\\
            e. Avg.\ Time Interval\\
            f. Longest Silence\\
            g. Code Standard\\
            h. Interaction Diversity\\
            i. Sentiment
        \end{tabular}
        &
        \begin{tabular}[t]{@{}l@{}}
            a. Attendance\\
            b. Attendance Skew\\
            c. Overall Meeting Time\\
            d. Task Fidelity\\
            e. Assignment Fidelity\\
            f. Task Diversity
        \end{tabular} \\
        \hline
    \end{tabular}
    \vspace{-5px}
    \caption{Lists of categorised metrics.}
\end{table}

\textbf{Submission}. We use version control logs to capture commit timing and volume. Following \cite{mitra2024analyzing}, we track net lines (adds-deletes) and both word and character counts to normalise for writing style. Temporal features include average inter-commit time and workload skew across the project timeline. Isolated quality is assessed via existing automated graders across multimodal contributions, supplemented by \textit{subjective} quality evaluation in the category below.

\textbf{Conversation}. We quantify interaction by message count; activity by response latency and silence; quality by sentiment/tone, length, and readability (i.e., structure, grammar, word choice); and diversity by how widely members interact to flag potential cliques.

\begin{figure}[htbp]
    \centering
    \includegraphics[width=0.7\textwidth]{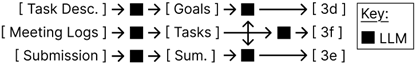}\vspace{-2px}
    \caption{Deriving abstract metrics via LLMs.}
    \label{fig:feature-adoption}
\end{figure}

\textbf{Coordination}. As displayed above, we use LLMs to extract goals, tasks and work summaries from the task description, meeting logs, and contributions to compute task fidelity (the relevance of meeting outcomes to the original project goals), task diversity (the spread of tasks), and assignment fidelity (whether assigned tasks were completed). Presence is gauged from attendance and meeting time, with skew for additional temporal context. 

\subsection*{[5.2] Objective Measures}

In this stage, we normalise and aggregate calculated metrics to produce objective measures of contribution, expanding the process with semantic and hypothetical embedding for stronger characterisations of more abstract notions of contribution.

\begin{figure}[htbp]
    \centering
    \includegraphics[width=0.95\textwidth]{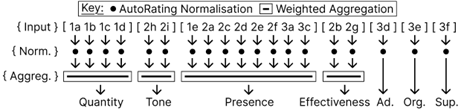}\vspace{-2px}
    \caption{Compiling metrics into base measures.}
    \label{fig:feature-adoption}
\end{figure}

We group the initial input metrics according to the category they align with in the conflict framework to derive objective metrics from collected evidence. In every category, each metric is normalized via AutoRating, and then aggregated with a weighted mask that sums to 1 (adjustable to emphasise different aspects of performance). This first aggregation produces \textit{base measures}, used to calculate markers of conflict.

\begin{figure}[htbp]
    \centering
    \includegraphics[width=0.95\textwidth]{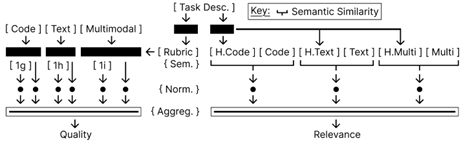}\vspace{-2px}
    \caption{Base measures with semantic and hypothetical embedding.}
    \label{fig:feature-adoption}
\end{figure}

AI graders improve with access to a rubric and examples of grading\cite{zhao2025language}. Thus, we produce an assessment guide to help with subjective quality evaluation performed by an AI grader for each contribution type. We pair this with deterministic tool grading for the total quality score. To determine relevance, we generate hypothetical documents that are expected to be completed to achieve the task outcomes. The similarity between these and the actual project output relays the utility of each student's contribution.  

\begin{figure}[htbp]
    \centering
    \includegraphics[width=1\textwidth]{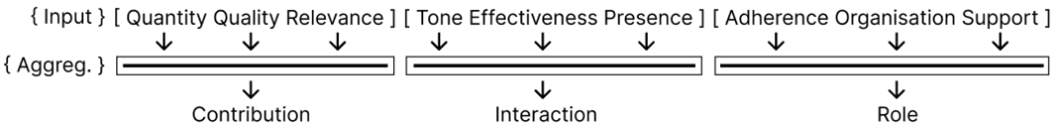}\vspace{-2px}
    \caption{Compiling objective measures from base measures.}
    \label{fig:feature-adoption}
\end{figure}

Finally, all the base measures are aggregated into three holistic \textit{objective} measures. We posit that having this select number of values prevents the instructor from being overwhelmed by statistics, whilst still conveying the performance of each student comprehensively.

\subsection*{[5.3] Conflict Markers}

We flag potential group conflicts by computing the Gini index for each base measure to quantify inequality in individual contributions\cite{hundhausen2023combining}. We propose two scenarios: (A) high Gini with an above average individual score entailing an isolated high performance; (B) low Gini with a below average score entailing an isolated low performance. We map these scenarios to measure-specific implications derived from the conflict conditions in our framework.

\begin{table}[h!]
\centering
\begin{tabularx}{\textwidth}{|p{0.10\textwidth}|p{0.265\textwidth}|p{0.265\textwidth}|p{0.2665\textwidth}|}
\hline
\textbf{Scenario} & \hspace{37px}\textbf{Quantity} & \hspace{42px}\textbf{Quality} & \hspace{35px}\textbf{Relevance} \\
\hline
\centerline{A} &
Overcentralisation in contribution. I.e., few members doing most of the work. &
A team member with stronger skills and knowledge putting in disproportionately larger effort. &
A stronger understanding of the task that is not shared across the team. \\
\hline

\centerline{B} &
Social loafing. I.e., few members not carrying their weight. &
Few students not matching the team competence standard. &
Contributed in a way that wasn’t as useful for the team.\\
\hline

\end{tabularx}
\vspace{-5px}
\caption{Conflict Markers (Contribution).}
\label{tab:equal_width2}
\end{table}
\begin{table}[h!]
\centering
\begin{tabularx}{\textwidth}{|p{0.10\textwidth}|p{0.265\textwidth}|p{0.265\textwidth}|p{0.2665\textwidth}|}
\hline
\textbf{Scenario} & \hspace{45px}\textbf{Tone} & \hspace{25px}\textbf{Effectiveness} & \hspace{37px}\textbf{Presence} \\
\hline
\centerline{A} &
  Professionalism and equal treatment in a team with cliques or negative communication. &
Good communicator that kept the team on track. &
High engagement in a more absent team. \\
\hline

\centerline{B} &
 Team member who behaved more negatively compared to the remainder of the team. &
Team members who communicated poorly compared to the rest of the team. &
 Uncooperative or absent team member. \\
\hline
\end{tabularx}
\vspace{-5px}
\caption{Conflict Markers (Interaction).}
\label{tab:equal_width2}
\end{table}
\begin{table}[h!]
\centering
\begin{tabularx}{\textwidth}{|p{0.10\textwidth}|p{0.265\textwidth}|p{0.265\textwidth}|p{0.2665\textwidth}|}
\hline
\textbf{Scenario} & \hspace{32px}\textbf{Adherence} & \hspace{26px}\textbf{Organisation} & \hspace{38px}\textbf{Support} \\
\hline
\centerline{A} &
Higher responsibility in a team with less stringent protocols for action. &
Taking the lead in structuring and coordination. &
Extra effort to fulfil their role and help the team. \\
\hline

\centerline{B} &
Failure to follow agreed-upon and established distribution of work. &
Lack of initiative and structure, failure to engage with organisational needs. &
Low level of soft contribution and engagement. \\
\hline

\end{tabularx}
\vspace{-5px}
\caption{Conflict Markers (Role).}
\label{tab:equal_width2}
\end{table}

\subsection*{[5.4] Expert Analysis}

We use hierarchical prompts with decomposed queries to infer the situation from input metrics, measures, and flagged conflict markers. We surface salient local features per category, and pair this with global information (i.e., objective measures) to form an overall conclusion that is screened with a double-pass validation. Thus, the instructor receives judgment backed by evidence-based reasoning and minimised hallucinations.

\begin{figure}[htbp]
    \centering
    \includegraphics[width=0.85\textwidth]{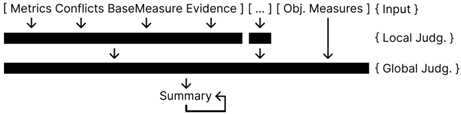}\vspace{-2px}
    \caption{Generating expert opinion via LLMs.}
    \label{fig:feature-adoption}
\end{figure}

This stage could benefit from further research into maximising LLM effectiveness (prompt engineering, Chain-of-Thought, etc.) to fine-tune judgments and improve summarisation. Further, validation passes and filtering should maintain regulatory compliance and an assistive stance. Human oversight during reasoning may also increase interpretability and inform precedence over different pieces of evidence.

\subsection*{[5.5] Accounting for Subjective and Contextual Data}

Additional data includes past grades, student history, and peer assessment feedback. These can either be numeric and directly translate towards metrics, or subjective, and thus harder to quantify. As seen below, past student grades can be paired with context (e.g., absence due to personal circumstances) as the estimated time relative to the project length, producing \textit{numeric} values that can be normalized and aggregated into an adjustment factor that directly affects the students’ final objective measures.

\begin{figure}[htbp]
    \centering
    \includegraphics[width=0.7\textwidth]{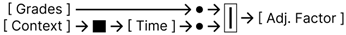}\vspace{-2px}
    \caption{Feauters in contribution assessment tools.}
    \label{fig:feature-adoption}
\end{figure}

For subjective data (e.g., peer assessment), we propose using classification to sort items into base measures, and appending to the original calculations. To account for survey bias, corrective normalisation can be employed\cite{li2001refinements}. 
Existing methods tackle trustworthiness in both types of data by gauging reviewer reliability; however, this is not directly applicable here. We urge for further research into investigation techniques accounting for gaming strategies and unreliable information in student feedback. Where honesty is uncertain, we suggest eschewing the subjective part of this system, and only incorporating the data in global judgment.

\subsection*{[5.6] Considerations}

\textbf{Limitations}. There are some cases where our system may show a notable decrease in accuracy, which we identify here: (1) Where evidence is missing, and does not fall under the outlined categories. (2) Where students use gaming strategies, and submit contradictory or misleading evidence. (3) Large cultural, linguistic, or personal differences. Additionally, the system must maintain an assistive/advisory stance and cannot provide direct judgment, which may indirectly limit assistance ability.

\textbf{Practical Challenges}. Implementation may pose some challenges: (1) The compatibility of data for parsing/exporting purposes. (2) Reconciling identity across multiple sources of contribution. (3) Judging quality of media contributions. (4) Identifying contributors for evidence. Nevertheless, we believe these issues won’t significantly affect feasibility.

\section*{[6] Policy Review}

The system necessarily (1) processes student data and assessment criteria, (2) applies algorithmic LA and AI-based methodology, and (3) evaluates relative engagement and contribution. Thus, our review centres on data governance, AI and analytical processing, and assessment performance management. We explore two main areas to do this: institutional policy, as well as statutory and regulatory compliance. Given that our team operates outside of the University of Warwick and WMG, our investigation will focus on UK policy, with brief coverage elsewhere.

\textbf{Data Governance}. Warwick dictates: “Warwick data should only be used in systems and tools provided to staff”\cite{WarwickAIEdTools2025} and that “it is not permissible to upload student work to an online platform without their consent”\cite{warwickAILearning2025}. Warwick’s IG02 Data Protection Policy\cite{WarwickDataProtectionIG02_2025} strongly mirrors Article 5 of the UK GDPR with data protection principles, as well as restrictions on data transfer location, and the ability for the data subject to exercise their rights. This is also the case for WMG\cite{WMGAcademyPolicies2025}, and indeed, all university policies should have policies that demonstrate compliance under GDPR\cite{JiscIntelligentCampus2023}. With this in mind, we defer to public law with regard to data protection – more specifically, article 5(1), that should “lie at the heart of [an] approach to processing personal data”\cite{icoDataProtectionPrinciples}.

\textbf{AI Policy}. Warwick states that “internal Warwick data” should not be given to an AI to train on\cite{warwickAITools2025}, nor supplied as a prompt\cite{warwickAILearning2025} without that student’s consent. Warwick’s Artificial Intelligence Information Compliance Policy\cite{Warwick_IMP02_2025} outlines the use of AI technologies – more specifically, machine learning (ML) and large language models (LLMs). They enforce data restrictions, such that certain data cannot be used by AI unless approved by the IDG or the Research Ethic Committee, on top of additional security principles about human involvement in quality assurance, and AI output risks.

The UK AI-Regulation White Paper\cite{DSIT_AI_Regulation_WhitePaper_2023} repeatedly highlights education as a domain where "high-impact outcomes" justify extra care, outlining five cross-sectoral principles. The DfE policy paper on generative artificial intelligence in education\cite{DfE_GenAI_Education_2023} mirrors this sentiment, calling for a "human-in-the-loop", and that any AI tool handling pupil data must meet UK GDPR.

The European Commission’s EU AI Act 2024/2025\cite{EuropeanParliament_EUAIAct_2023} establishes transparency requirements and risk qualifications for AI systems, and more relevantly, “AI systems intended to be used to evaluate learning outcomes in educational institutions of all levels”\cite{EUAIAct_AnnexIII_2025}. UNESCO’s report on GenAI in education and research charts guidelines for policy, and discusses ethical issues like equity, psychological impact, hidden bias, and discrimination\cite{Miao_Holmes_UNESCO_GenAI_2023}.

\textbf{Assessment Regulations}. In the case of automated decision-making, which precludes human involvement, the student can object to its use, and request human review\cite{WMGPrivacyStudents2024}\cite{EU_AI_Act_2025}. As such, where a tool takes an advisory role and avoids making judgements itself, it instead qualifies as profiling: "automated processing of personal data ... to evaluate certain personal aspects relating to a natural person"\cite{icoAutomatedDecisionProfiling} – such as academic performance. This has precedent at the University of Leicester, for instance, where its policy on AI states "AI may have a role in generating marks and feedback, however the marks and feedback for each student will be reviewed by the relevant expert member of staff."\cite{UoL_AI_Policy_2024} With regards to data collected for the purpose of assessment, using it does not require ethical approval\cite{warwickEthicalApproval}. However, it may require a Data Protection Impact Assessment (DPIA)\cite{icoDPIA}. 

\section*{[7] Conclusion}

Regarding a tool designed to investigate conflict in collaborative tasks and teamwork using AI, we identify a precedent and argue for feasibility under current policy, so long as care is taken about special category data, the correct protocols are followed (e.g., DPIA, IDG), and judgment is advisory, transparent, and ethical. 
	
We identify common features in existing tools (normalisation, Likert-scale surveys, reviewer reliability, early warning systems, AI screening, data visualization, and LMS integration), which we urge developers consider for a tool to be competitive, and note a gap in AI integration and conflict resolution techniques.

Our approach to automated conflict identification is more consistent and homogeneous compared to existing tools, which combine multiple calculations. Unlike previous work, we base our conflict analysis on an organised set of issues sourced from all aspects of contribution, taking inspiration from existing instruments.

Our architecture enhances reliability with judgments backed by metrics, combines subjective and objective evidence to increase accuracy, and mitigates bias with normalisation and correction. We use AI to capture abstract notions of contribution, quantify student context, and perform reasoning. Overall, we hope this AI-driven interpretation of results diminishes the inherent distrust in technocratic solutions, offers value to investigations, and decreases administrative load.

Future work could adopt this architecture, with possible studies into improving inter-rater reliability with heterogeneous agents, and mitigating fairness issues stemming from LLM bias. Enhancements may target trustworthiness, adversarial robustness, and transparency. Finally, one could address the research gap in how instructors can engage with model-generated advice and conduct AI-assisted investigation – beyond keeping the “human-in-the-loop”.

\newpage
\bibliographystyle{abbrv}
\bibliography{bibliography}

\end{document}